\documentclass[journal]{IEEEtran}
\ifCLASSINFOpdf
  \usepackage[pdftex]{graphicx}
\else
\fi

\usepackage{amsmath}
\usepackage{amsfonts}
\usepackage[table,dvipsnames]{xcolor}

\usepackage[hyphens]{url} 
\usepackage{lineno,hyperref}
\hypersetup{colorlinks=true,urlcolor=MidnightBlue,linkcolor=MidnightBlue,citecolor=MidnightBlue}

\usepackage{multirow}

\usepackage{filecontents}

\begin{document}
%
\title{Sequence Mining and Pattern Analysis in Drilling Reports with Deep Natural Language Processing}

\author{J\'ulio~Hoffimann,~\IEEEmembership{}
        Youli~Mao,~\IEEEmembership{}
        Avinash~Wesley,~\IEEEmembership{}
        and~Aimee~Taylor~\IEEEmembership{}
\thanks{J. Hoffimann is with the Department
of Energy Resources Engineering, Stanford University, Palo Alto, CA, 94305 USA e-mail: juliohm@stanford.edu.}
\thanks{Y. Mao, A. Wesley and A. Taylor are with Halliburton, Houston, TX, 77032 USA e-mail: e-mail: youli.mao@halliburton.com, avinash.wesley@halliburton.com, aimee.taylor@halliburton.com.}
\thanks{}}

\markboth{}%
{Hoffimann \MakeLowercase{\textit{et al.}}: Deep NLP for Sequence Mining and Pattern Analysis in Drilling Reports}
%



\maketitle

\begin{abstract}
Drilling activities in the oil and gas industry have been reported over decades for thousands of wells on a daily basis, yet the analysis of this text at large-scale for information retrieval, sequence mining, and pattern analysis is very challenging. Drilling reports contain interpretations written by drillers from noting measurements in downhole sensors and surface equipment, and can be used for operation optimization and accident mitigation. In this initial work, a methodology is proposed for automatic classification of sentences written in drilling reports into three relevant labels (EVENT, SYMPTOM and ACTION) for hundreds of wells in an actual field. Some of the main challenges in the text corpus were overcome, which include the high frequency of technical symbols, mistyping/abbreviation of technical terms, and the presence of incomplete sentences in the drilling reports. We obtain state-of-the-art classification accuracy within this technical language and illustrate advanced queries enabled by the tool.
\end{abstract}


%
\IEEEpeerreviewmaketitle

\section{Introduction}

\IEEEPARstart{D}{rilling} activities in the oil and gas industry are a shared concern among energy companies, government agencies, and the general public because they can affect both company profitability and the natural environment. Having these activities fully reported and classified is challenging to a limited human workforce; ideally, this task would be assigned to an intelligent reporting system during the drilling operation in real time.

The energy industry is in its initial steps toward the ultimate goal of smart reporting. In their paper, ``Augmenting Operations Monitoring by Mining Unstructured Drilling Reports,'' Sidahmed et al. \cite{Sidahmed_2015} apply traditional natural language processing (NLP) techniques to drilling reports collected daily from three wells and show that, although unstructured, the text can provide new insights about drilling events (e.g., stuck pipe) that are not easily captured with other types of data. The work of Sidahmed et al. is one of the few published articles on the subject and serves as a proof of concept that rich information in the form of text is being underused by the industry. Although very enlightening, the paper does not benefit from the recent advances in NLP and deep learning, which can greatly scale information retrieval to complex industrial settings.

Researchers with companies and universities, such as Google, Facebook, and Stanford, have devoted a great deal of time considering how to process text for information retrieval \cite{Sanderson_2012,Manning_2008,Buttcher_2010}. From the most basic task of text classification to more elaborated summarization and question answering software, researchers have developed novel algorithms capable of learning semantics with limited or no human supervision. Such learning process depends on the availability of large datasets in which deep neural networks are trained on various NLP tasks.

In this work, recent developments in deep natural language processing (or deep NLP) are applied to automatically classify sentences in thousands of drilling reports and to identify companies' behavior. This tool can be used offline by an energy company interested in verifying old drilling reports for operation patterns or by a government agency interested in investigating the aftermath of environmental disasters. In the future, this tool could be used in real time to enhance decision support systems, to help mitigate drilling costs associated with non-productive time, and to reduce the risk of accidents.

This paper is organized as follows. \autoref{sec:literature} reviews state-of-the-art deep natural language processing models for text classification. \autoref{sec:methodology} introduces the dataset and methodology used. \autoref{sec:results} presents the results obtained with drilling reports and discusses the accuracy of the methods. \autoref{sec:conclusion} describes conclusions about the applicability of the proposed tool and points to future directions.

\section{Literature review}\label{sec:literature}

Natural language understanding and modeling has been a subject of research in artificial intelligence since before the 1980s. In his article, ``n-Gram Statistics for Natural Language Understanding,'' Suen \cite{Suen_1979} correctly predicts the success of machine intelligence in NLP and presents classical n-gram statistics of the English language based on a corpus of 1 million words. Although more (memory-) efficient than vocabulary-based methods, the n-gram statistics that were very powerful at the time are no longer the most appropriate model to manage the massive amount of text generated daily on the web (e.g., Google corpus, Wikipedia corpus) \cite{Bengio_2000,Schwenk_2007,Mikolov_2011a}, nor can they easily capture hidden semantic relationships needed for more challenging tasks, such as information retrieval, summarization and question answering \cite{Manning_2008,Tellex_2003}.

Mikolov et al. \cite{Mikolov_2013} introduce a vector encoding for text words that outperforms traditional NLP techniques based on co-occurrence counts and n-gram statistics. The encoding effectively captures word semantics and shows surprisingly linear relationships (e.g., ``man'' - ``woman'' + ``queen'' $\approx$ ``king'') when trained on a sufficiently large corpus. These learned relationships and word similarities are superior to those derived with traditional minimum edit distance \cite{Navarro_2001}, and can be exploited to achieve much deeper inference, as in the astonishing example ``Russian'' + ``river'' $\approx$ ``Volga river'' \cite{Mikolov_2013}.

The continuous encoding proposed by Mikolov et al. is obtained by means of a variation of cross-entropy minimization. To efficiently obtain an optimal language model, the authors introduce a proxy logistic regression that maximizes the probability of word pairs in the same context and minimizes the probability of word pairs drawn at random from the unigram distribution. This technique is called negative sampling, which is a particular case of noise-contrastive estimation (NCE) of partition functions \cite{Gutmann_2012}.

Two variants of the algorithm exist, depending on what is predicted and what is given in the conditional probabilities: the continuous bag-of-words (CBOW) and the skip-gram models. The CBOW model is trained to predict the center (or target) word in the context window, whereas the skip-gram model predicts context words given the center word in the window. The choice between these two models is usually made by considering that the latter is better suited for learning representations of infrequent words in the corpus.

In their article, ``Glove: Global Vectors for Word Representation'', Pennington et al. \cite{Pennington_2014} discuss the properties that an encoding algorithm must obey for word vectors to display semantic relationships similar to those introduced by Mikolov et al. The authors propose a slightly different objective function, inspired by simple experiments with probability ratios, and demonstrate superior accuracy in word analogy tasks. These accuracy improvements, however, are rarely manifested in practice with other NLP tasks.

In deep NLP, trained word vectors are input features to non-linear classifiers/regressors. These vectors represent weights in a deep neural network trained for a particular task \cite{Kalchbrenner_2014,Irsoy_2014,Kim_2014}. The overall accuracy of deep learning methods in NLP depends on the neural network architecture and its capability of memorizing context information.

Deep network architectures have been successfully applied to different areas of computer vision, such as face detection and recognition \cite{Rowley_1998,Lawrence_1997}, human action recognition \cite{Ji_2013}, and handwriting recognition \cite{Graves_2009,Frinken_2012}. Mostly convolutional, these networks compose input features (or words) into high-level signals (or sentences), and can be exhaustively trained for solving challenging tasks, such as dense captioning \cite{Johnson_2016}.

Recurrent neural network (RNN) architectures have been consistently investigated in deep NLP because of their resemblance to linguistic grammars \cite{Jozefowicz_2015,Irsoy_2014,Sutskever_2011} and superior classification accuracy in language-related tasks. RNN architectures with memory cells have been developed that are capable of retaining long sequence patterns. As the most prominent member of this category, long short-term memory (LSTM) networks are suitable for text classification and language comprehension \cite{Hochreiter_1997}.

Recently, deep neural networks have gained attention in NLP because of their state-of-the-art accuracy, as compared to systems carefully designed for specific languages and tasks \cite{Mikolov_2011b}. Their adoption in information retrieval systems of technical endeavor, such as the present work, has just started.

\section{Methodology}\label{sec:methodology}

This section begins with an introduction to the dataset from an actual petroleum field. Basic summary statistics and traditional NLP views of the data are presented to illustrate the challenges with this technical language. This section also includes a detailed description of the workflow with other modeling decisions.

\subsection{Data}

An energy company provided 9670 daily reports written by drilling engineers in a field with 303 oil wells. These reports can be classified into two groups according to the drilling state: productive and non-productive time. A drilling operation is considered to be in non-productive time (NPT) when actions are taken to solve an issue or accident. For example, drilling tools can sometimes become stuck as a result of miscalculations and limited understanding about the subsurface. In this case, the NPT includes the fishing (or rescue) of the tool, as well as other minor adjustments before drilling resumes.

\begin{figure}[!t]
\centering
\includegraphics[width=3.5in]{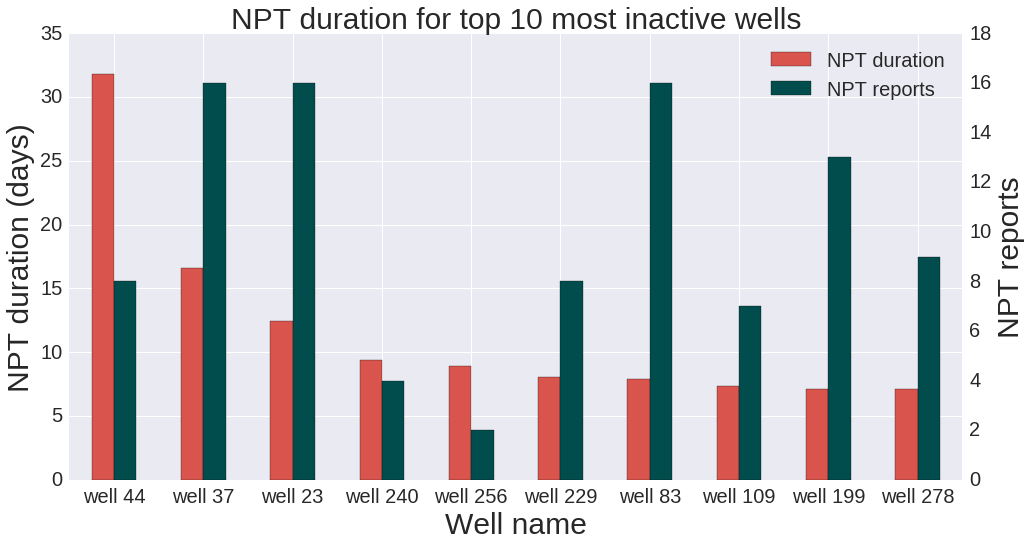}
\caption{NPT duration (red) and number of reports (green) for the top 10 most inactive wells in the field.}
\label{fig:npt_duration}
\end{figure}

In this study, 112 out of 303 wells contain NPT reports. The percentage $1 - \frac{112}{303} \approx 63\%$ is a rough estimate of the drilling operations performance, which is considered to be low. \autoref{fig:npt_duration} illustrates the total NPT duration and number of reports for the top 10 most inactive wells in the field.

\begin{figure}[!t]
\centering
\includegraphics[width=3.5in]{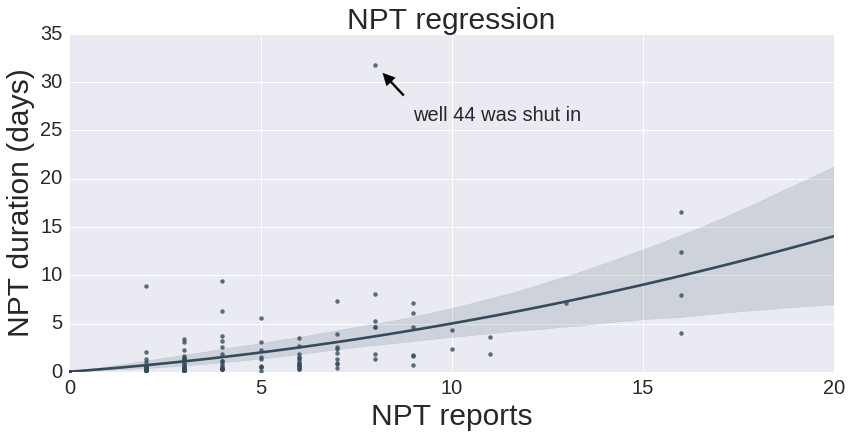}
\caption{Simple regression model showing expected positive correlation.}
\label{fig:regression}
\end{figure}

In \autoref{fig:npt_duration}, well 44 was shut in as a result of unusually high pressure. This action resulted in a multi-million dollar cost to the company and explains the small number of NPT reports for this well. Well 83 provides an example of good reporting practices given its short NPT duration. In general, a positive correlation is expected between the number of reports and NPT duration, as depicted with a simple regression model in \autoref{fig:regression}.

The remaining 303 - 112 = 191 wells in the field contain thousands of productive time (PT) daily reports with information that include equipment inspection and physical measurements at depth.

\begin{figure}[!t]
\centering
\includegraphics[width=3.5in]{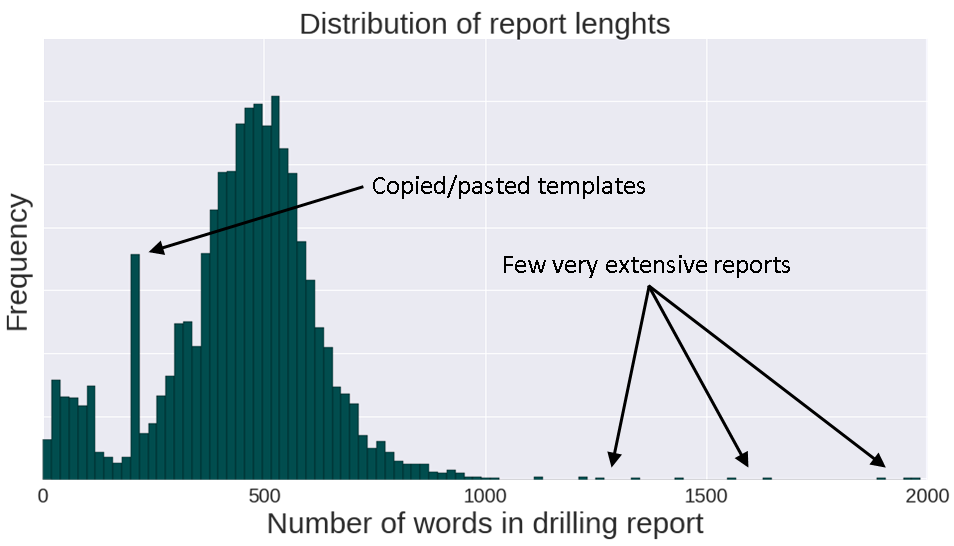}
\caption{Distribution of reports lengths measured by number of words.}
\label{fig:report_length}
\end{figure}

\autoref{fig:report_length} illustrates the distribution of report lengths for both PT and NPT calculated as the number of words in the text. At least two modes are observed in the length distribution: the first mode at approximately 100 words is associated with reports that only contain a summary paragraph; the second mode at approximately 500 words is associated with reports that contain additional remarks after a summary.

\begin{figure}[!t]
\centering
\includegraphics[width=3.5in]{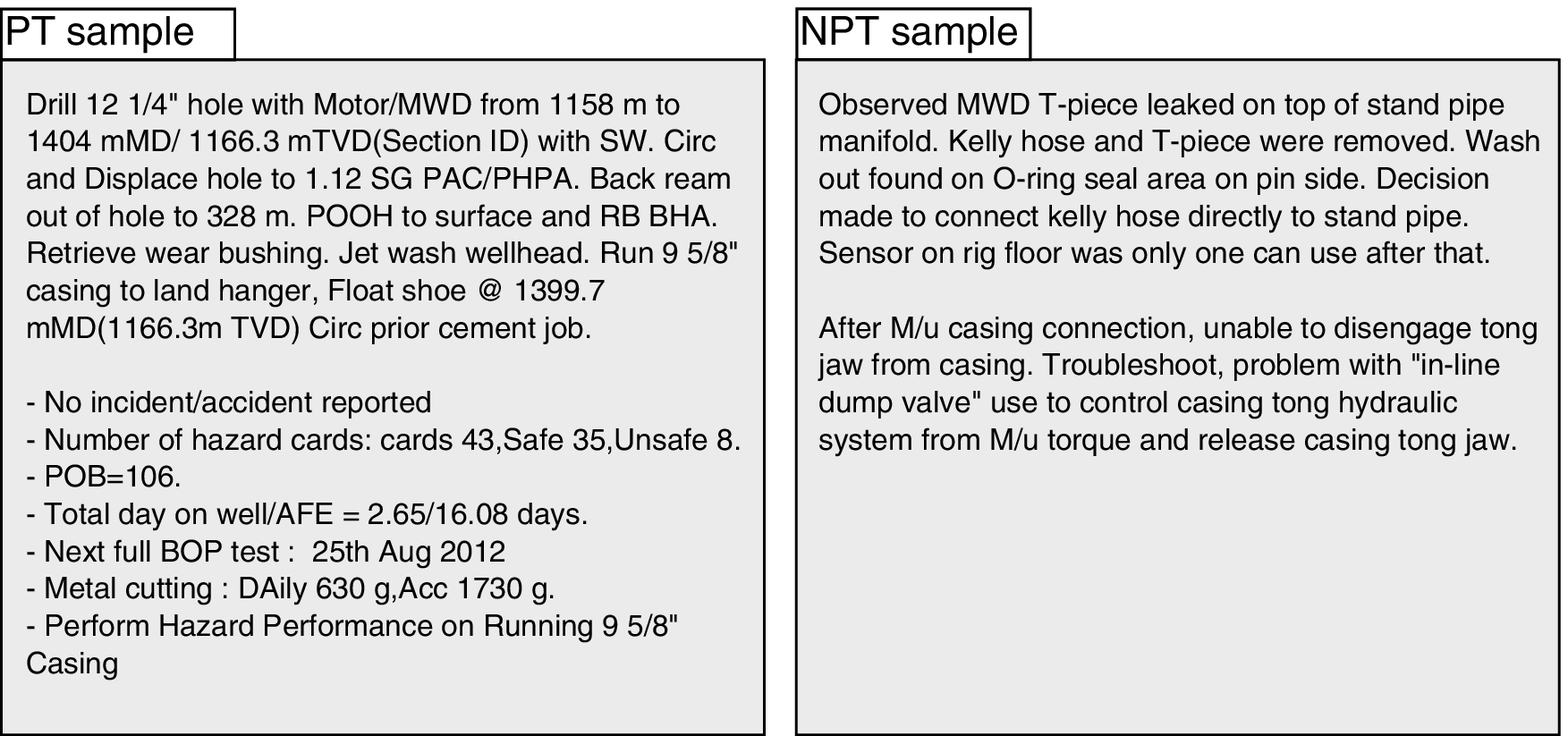}
\caption{Samples of PT and NPT reports illustrating their different format.}
\label{fig:PT_VS_NPT}
\end{figure}

PT and NPT reports differ in their format, as shown by the samples in \autoref{fig:PT_VS_NPT}. PT reports contain a short summary of the activity, followed by technical remarks. NPT reports usually consist of a long summary that describes the accident or issue in the drilling operation. Despite the format, these two types of reports share context and therefore meaningful relationships.

\begin{figure}[!t]
\centering
\includegraphics[width=3.5in]{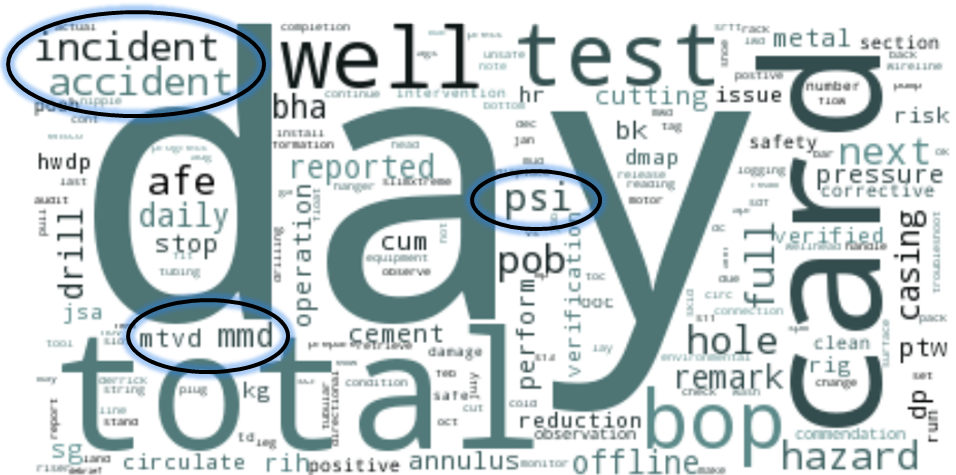}
\caption{Word cloud for all drilling reports showing high frequency of physical units, acronyms, and abbreviations. The words ``incident'' and ``accident'' are consistently reported.}
\label{fig:word_cloud}
\end{figure}

\begin{figure}[!t]
\centering
\includegraphics[width=3.5in]{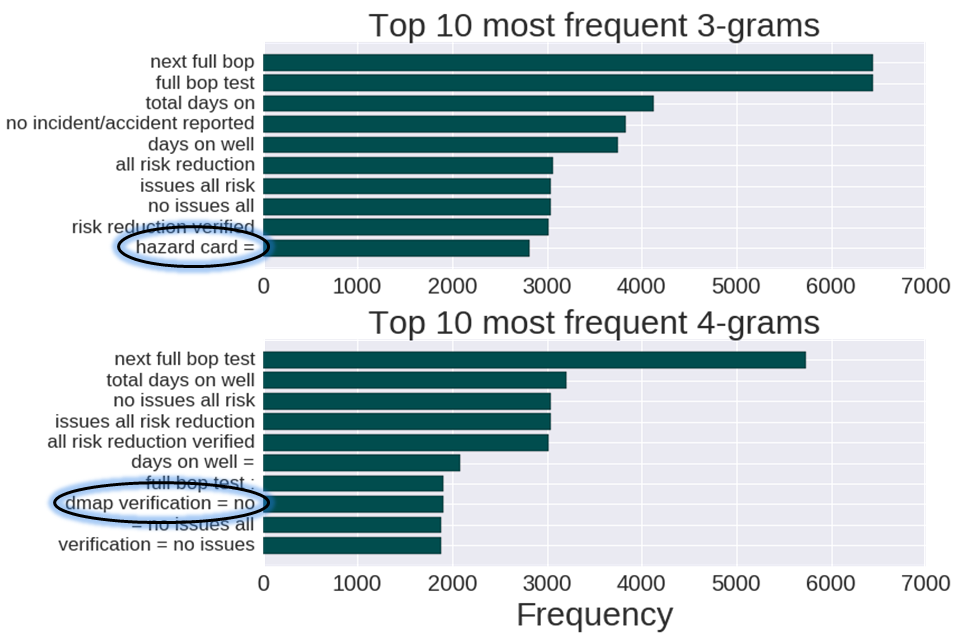}
\caption{Top 10 most frequent 3-grams and 4-grams showing unstructured sentences as the result of shortcuts adopted by drillers in their day-to-day reporting activities.}
\label{fig:ngram}
\end{figure}

\autoref{fig:word_cloud} illustrates the word frequency cloud for all reports. The total number of words (or tokens) is 859,361, and the vocabulary size is 22,337. Several units, such as ``psi'', ``mTVD'' and ``mMD'' are quite frequent in the language, and target words, such as ``incident'' and ``accident'' are consistently reported. \autoref{fig:ngram} shows the top 10 most frequent 3-grams and 4-grams in the corpus. Sentences that contain shortcuts adopted by drillers in their daily reports are highlighted.

Sentences extracted from NPT reports are labeled as EVENT, SYMPTOM, or ACTION by a drilling expert. An event is defined as a major accident or failure in the operation. For example, the drilling tool may become stuck in the pipe. The driller can either take various actions after an event has happened, or be proactive and modify the drilling plan before failure, based on observable symptoms. Examples of symptoms include erratic torque during drilling, fluid leakage, and over pressure. Inevitably, the labeling is unbalanced with 28\% EVENT, 15\% SYMPTOM, and 57\% ACTION, which reflects that events and symptoms are usually followed by multiple actions during drilling. It also reflects companies' behavior in reporting operations. In this field, 20 companies co-operate in the drilling cycle.

\subsection{Methods}

\begin{figure}[!t]
\centering
\includegraphics[width=3.5in]{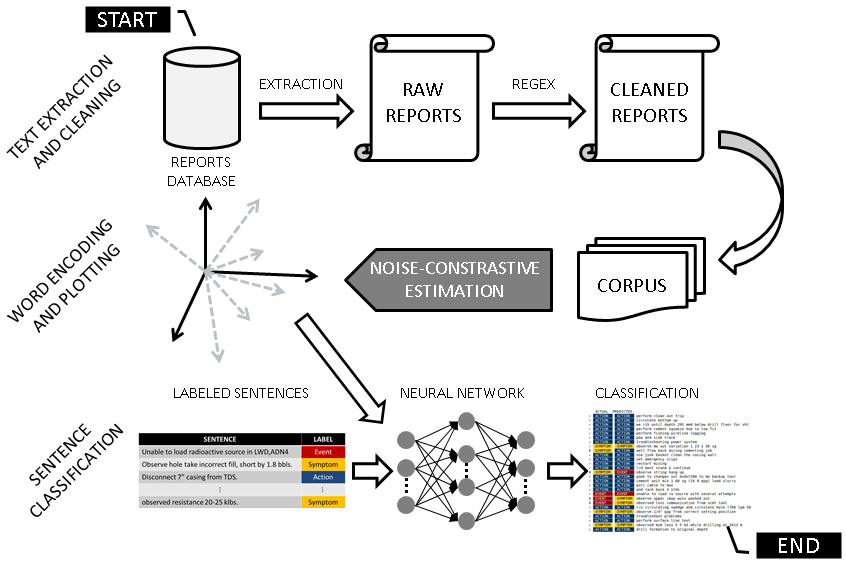}
\caption{Workflow for classification of sentences in drilling reports.}
\label{fig:workflow}
\end{figure}

\autoref{fig:workflow} illustrates the workflow for classification of sentences in drilling reports. The process begins by extracting the daily operational notes from the database of 303 wells and by concatenating the text in chronological order. Next, the reports are cleaned with the regular expressions listed in \autoref{tab:regex}. These expressions purge symbols that are considered to be meaningless in the context in which they appear, and can be interpreted as a denoising layer. After cleaning, the total number of tokens and the vocabulary size in the corpus are reduced to T = 810,375 and V = 17,623, respectively.

\begin{table}[!t]
\renewcommand{\arraystretch}{1.3}
\caption{Regular Expression Substitutions in Order of Execution. (Python syntax)}
\label{tab:regex}
\centering
\rowcolors{2}{gray!10}{gray!10}
\begin{tabular}{ccc}
\rowcolor{gray!30}
FROM & TO & PURPOSE \\
\verb!',\s'! & \verb!' '! & Commas at end of words \\
\verb!',([a-zA-Z])'! & \verb!' \1'! & Commas at end of words \\
\verb!'\((.*?)\)'! & \verb!' \1 '! & Enclosing parenthesis \\
\verb!'\xe2\x80\xa2'! & \verb!' '! & Bullet marks \\
\verb!'-\s'! & \verb!' '! & Dashes \\
\verb!'==+|\*\*+'! & \verb!' '! & Horizontal bars \\
\verb!'\[(.*?)\]'! & ' \verb! \1 '! & Enclosing brackets \\
\verb!'#|;'! & \verb!' '! & Pounds and semicolons \\
\verb!'_'! & \verb!' '! & Underscores \\
\verb!'\s/\s'! & \verb!' '! & Orphan forward slashes
\end{tabular}
\end{table}

\begin{figure}[!t]
\centering
\includegraphics[width=3.5in]{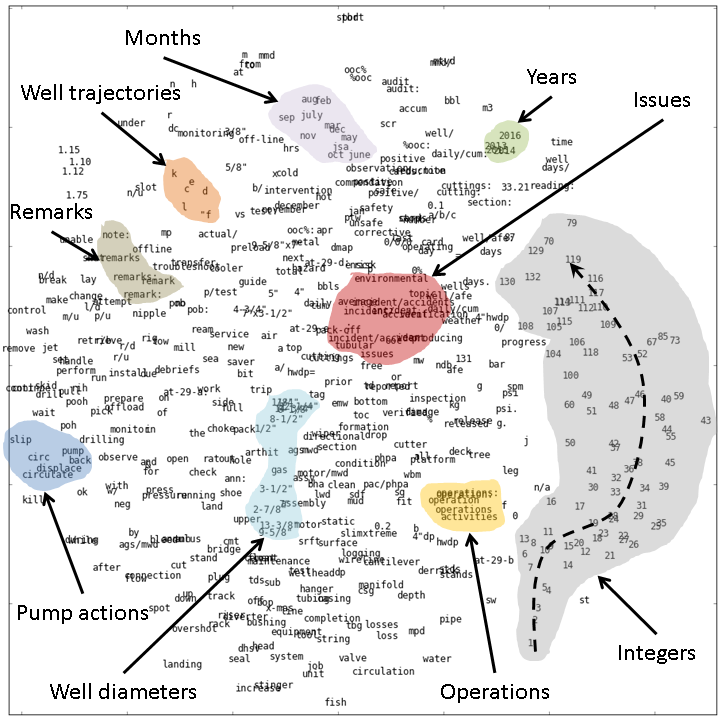}
\caption{t-SNE projection of top 500 most frequent words showing concepts like ``integers'', ``issues'', and ``well trajectories''.}
\label{fig:word_embedding}
\end{figure}

The Mikolov et al. methodology is strictly followed. The corpus is scanned with a fixed window of size $m=3$, and each word $w_i,\ i=1,2,\ldots,V$ in the vocabulary is assigned two random vectors $u_i,v_i \in [-1,1]^d$ with $d=300$ the embedding dimension. The word $w_i$ can either be in the center of a window, in which case $v_i$ is the associated vector representation, or an outer (or target) word for which $u_i$ is looked up likewise. Within a context window centered at a word $w_c$, a correct outer word $w_o$ is sampled. In addition, $k=64$ words $w_1,w_2,\ldots,w_k$ are sampled from the vocabulary at random from the unigram distribution $P(w)$. The probability of the pair $(w_c,w_o)$ is maximized and the probability of the pairs $(w_c,w_i),\ i=1,2,\ldots,k$ is minimized, simultaneously, with the objective function:
\begin{equation}
J_t = \log{\sigma(u_o^\top v_c)} + \sum_{i=1}^k \mathbb{E}_{w_i \sim P(w)}\left[\log{\sigma(-u_i^\top v_c)}\right]
\end{equation}
with $\sigma(x) = \frac{1}{1+e^{-x}}$ the sigmoid function. This process is repeated for all context windows throughout the corpus leading to the total objective $J = \sum_{t=1}^T J_t$. A batch of $b=128$ word pairs is processed at a time during minimization. Word vectors are updated iteratively with stochastic gradient descent and a learning rate of $lr=1.0$. At the end of the minimization, the average loss is approximately $2.02$, and the resulting embedding is illustrated in \autoref{fig:word_embedding} by means of a t-SNE projection \cite{Maaten_2008}. All hyper-parameters in the model were tuned by trial and error.

Vectors learned in this unsupervised stage cluster into concepts. As shown in \autoref{fig:word_embedding}, words such as ``incident'', ``accident'', ``issues'', and ``environmental'' appear together. Interesting patterns were observed, such as integers grouped in ascending order. The word embedding is robust to noise, as shown by the proximity of the words ``remarks'', ``remark'', ``remarks:'' Abbreviations are also captured as in ``circulate'' and ``circ''. All of these properties are extremely relevant for overcoming nuances of technical languages.

To classify sentences in drilling reports, three neural network architectures were tested: simple network with arithmetic averaging, convolutional neural network (CNN), and long short-term memory network (LSTM).

\begin{figure}[!t]
\centering
\includegraphics[width=3in]{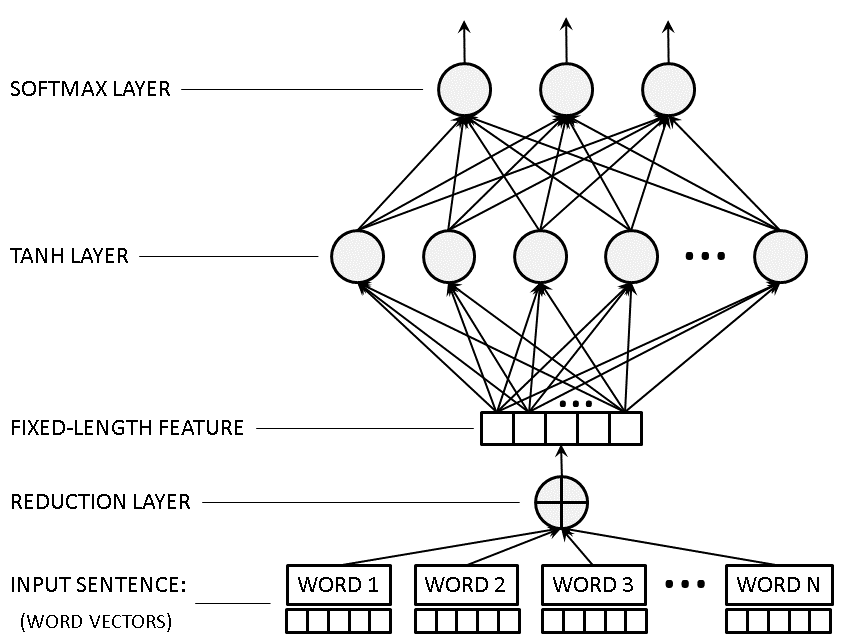}
\caption{Simple network with arithmetic averaging.}
\label{fig:simple_net}
\end{figure}

\paragraph{Simple network with arithmetic averaging} In this architecture, fixed-length features are assigned to sentences by averaging (i.e. reduction operation) their constituent word vectors. This feature is then passed to a fully connected hidden layer with 20 \emph{tanh} neurons followed by a \emph{softmax} classification layer. Words that are not present in the vocabulary are assigned the zero vector. The architecture is illustrated in \autoref{fig:simple_net}.

\begin{figure}[!t]
\centering
\includegraphics[width=3.5in]{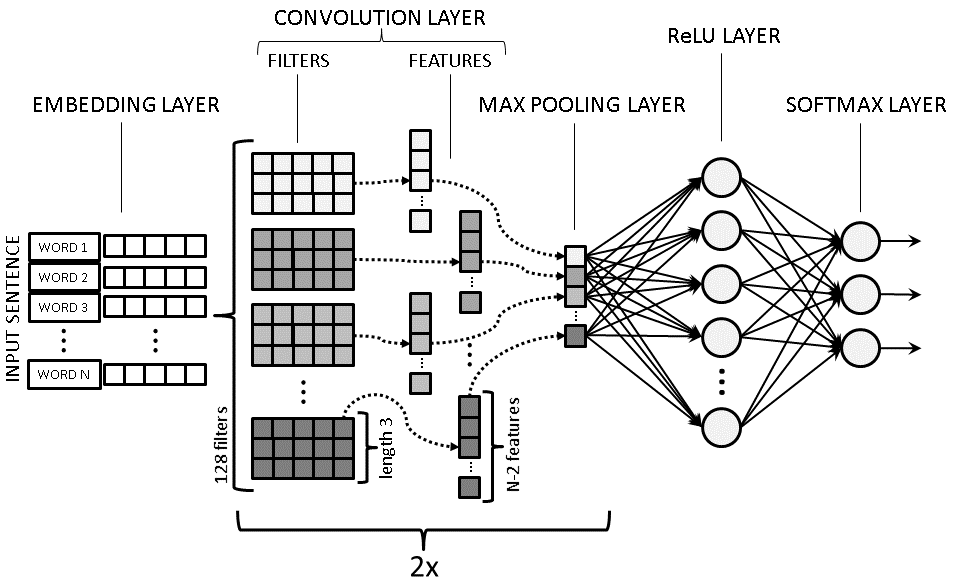}
\caption{Convolutional neural network.}
\label{fig:CNN}
\end{figure}

\paragraph{Convolutional neural network} Input sentences are padded to have at least the maximum sentence length in all drilling reports. The padding consists of introducing a special padding token not present in the corpus. In this architecture, pre-trained word vectors are passed to an embedding layer, which converts words in the vocabulary into the corresponding word vectors. Next, convolution and max pooling layers are interleaved twice in a total of four additional layers. The two convolutional layers consist of 128 filters of length 3 and the two max pooling layers halve their inputs. The architecture ends with a fully connected layer composed of 128 \emph{ReLU} neurons and a fully connected \emph{softmax} layer, as shown in \autoref{fig:CNN}.

\begin{figure}[!t]
\centering
\includegraphics[width=3.5in]{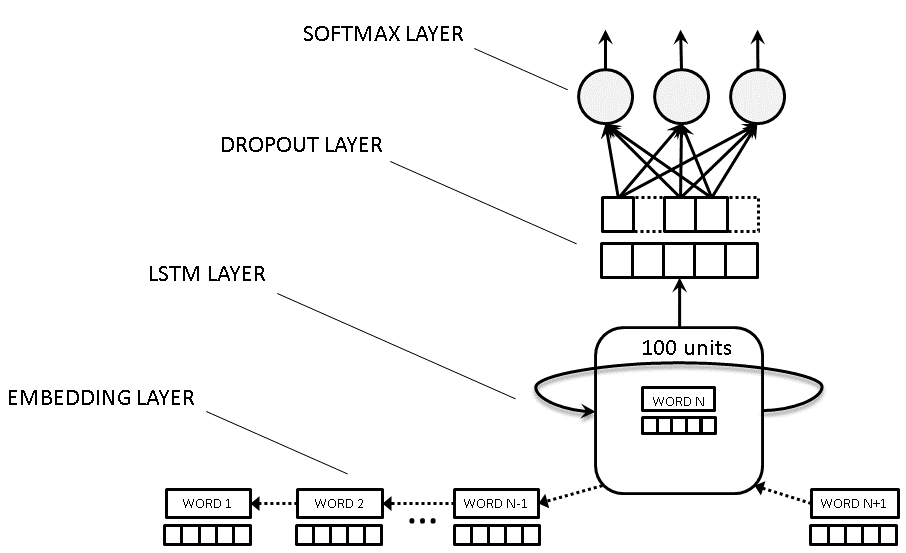}
\caption{Long short-term memory network.}
\label{fig:LSTM}
\end{figure}

\paragraph{Long short-term memory network} Similar to the convolutional neural network, this architecture begins with an embedding layer on padded input sentences. A long short-term memory layer with 100 neurons is appended, followed by 0.5 dropout. The architecture ends with a fully connected \emph{softmax} layer, as shown in \autoref{fig:LSTM}.

The neural networks are trained on a set of labeled sentences provided by a drilling expert. These sentences were extracted from NPT reports only. Each labeled sentence is then preprocessed with the same regular expressions used for cleaning the entire corpus; 80\% of them are used for training and 20\% are saved for testing.

\section{Results and discussion}\label{sec:results}

In this section, sentences that are present in drilling reports are classified as EVENT, SYMPTOM, or ACTION. Classification accuracy with 5-fold cross-validation is presented in \autoref{tab:results} for different neural network architectures. The results confirm that LSTM is superior in the task, followed by CNN and simple network with arithmetic averaging. The superior accuracy of LSTM is explained by its capability of memorizing context information in the input text, and is consistent with the deep NLP literature.

\begin{figure}[!t]
\centering
\includegraphics[width=3.5in]{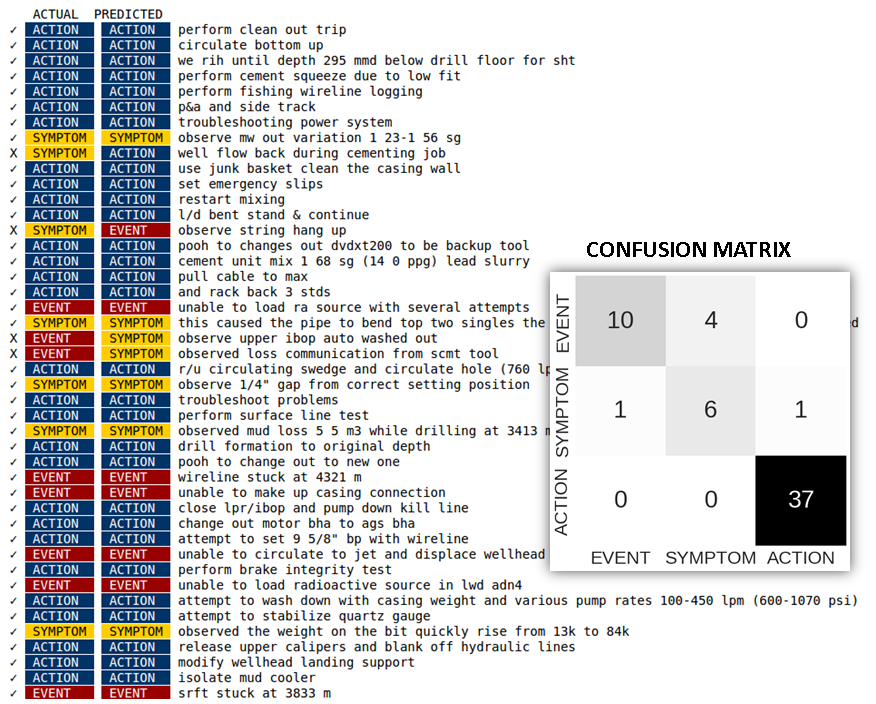}
\caption{Classification of a few unseen sentences extracted from the test set and corresponding confusion matrix.}
\label{fig:results}
\end{figure}

In \autoref{fig:results}, classification of unseen sentences extracted from the test set are shown along with the corresponding LSTM confusion matrix, which is a better evaluation metric than accuracy for classification tasks with unbalanced labels. Most misclassifications occur between symptoms and events, which corroborates that the training set contains few examples of these classes.

\begin{table}[!t]
\renewcommand{\arraystretch}{1.3}
\caption{Sentence Classification Accuracy.}
\label{tab:results}
\centering
\rowcolors{2}{gray!10}{gray!10}
\begin{tabular}{cc}
\rowcolor{gray!30}
Neural network architecture & \begin{tabular}{c}Mean accuracy \\ (5-fold cross-validation)\end{tabular} \\
Simple network with arithmetic averaging & 76.0\% \\
Convolutional neural network & 77.3\% \\
Long short-term memory network & 82.7\%
\end{tabular}
\end{table}

\begin{figure}[!t]
\centering
\includegraphics[width=3.5in]{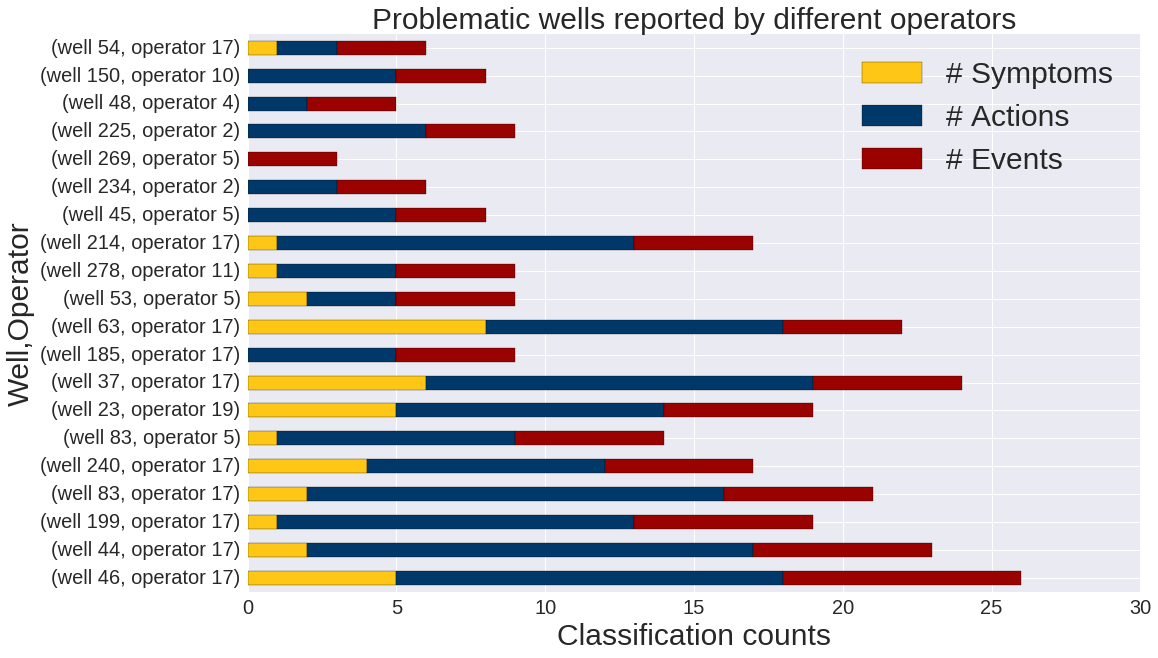}
\caption{Selective extraction of most problematic wells per reporting operator based on proposed classification.}
\label{fig:extraction}
\end{figure}

After training the LSTM network and testing it on the unseen labeled sentences, the next step consists of feeding the network with all sentences in NPT reports for the 303 wells in the field. The classification enables queries, such as the example shown in \autoref{fig:extraction}. In this plot, the most problematic wells are selected according to the number of sentences classified as events. In practice and more interestingly, the company that owns the field or an agency investigating an accident can retrieve all wells for which a specific symptom is followed by a specific action. The success rate of a remediation action can be estimated, for example, by analyzing the various outcomes in the classified sequence.

\begin{figure}[!t]
\centering
\includegraphics[width=3in]{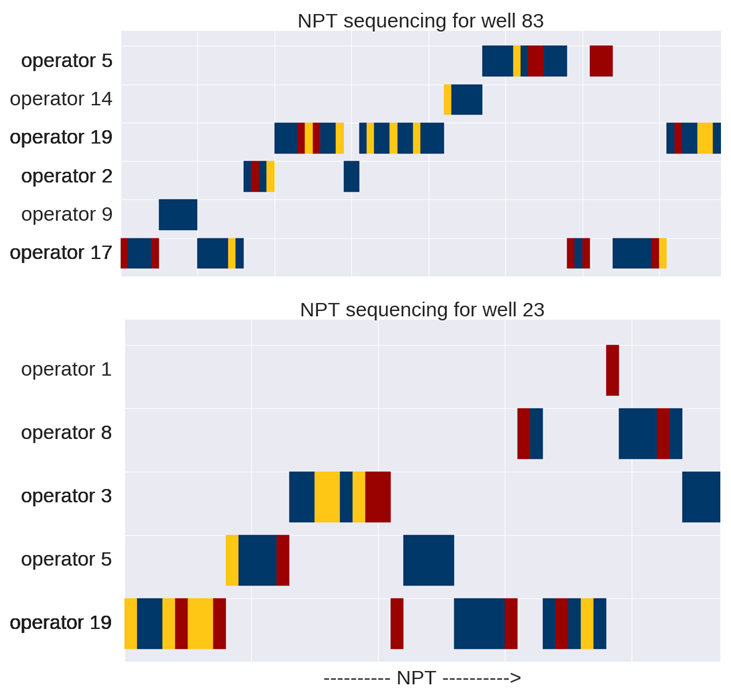}
\caption{NPT sequencing for wells 83 and 23. Operators show different reporting behavior according to their interests and roles.}
\label{fig:sequencing}
\end{figure}

In \autoref{fig:sequencing}, the NPT of two problematic wells is classified. Different companies were confirmed to have different interests and roles when reporting the same well. For example, on a normalized scale, the operator 19 reports less actions than the other operators on the same well, which corroborates internal business decisions. These illustrated differences in reporting behavior are also affected by the drilling stage, depth at which reports were written, and well location.

\section{Conclusion}\label{sec:conclusion}

In this work, a methodology was presented for information retrieval in drilling reports with deep natural language processing. The methodology was tested with 9670 reports from 303 wells in an actual field, and promising NPT sequencing results were obtained.

The skip-gram model with noise-contrastive estimation is capable of embedding words from drilling reports successfully. The algorithm produces good encodings, even with a small corpus, because of the repetitiveness of sentences in the reports and the limited vocabulary of the language.

Sentence classification can be performed accurately while drilling. Among all neural network architectures used in this work, LSTM is optimal for the task given its ability to memorize context information. More effort is required from energy companies, however, to refine the labeling and to begin considering a collaborative project for corpus enlargement.

Having drilling reports classified automatically for thousands of wells can help to mitigate accidents and improve our understanding of the sequences of actions taken by drilling companies. The methodology can be used to establish connections between classified text and other well data (e.g., well logs, rock samples), and can be applied with little or no modification to other fields that have a comparable number of technical reports and terms.



\section*{Acknowledgment}

The authors would like to thank the energy company that provided the data for this study and the many developers leading the open source projects TensorFlow\footnote{\url{https://www.tensorflow.org}} and Keras\footnote{\url{https://keras.io}}, both used for the implementation of this work.

\ifCLASSOPTIONcaptionsoff
  \newpage
\fi



\bibliographystyle{IEEEtran}
\bibliography{references}



\begin{IEEEbiography}[{\includegraphics[width=1in,height=1.25in,clip,keepaspectratio]{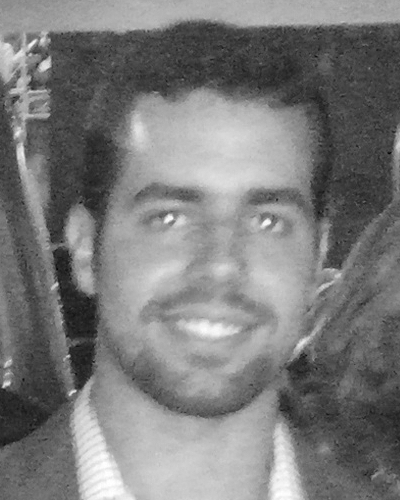}}]{J\'ulio Hoffimann}
received a B.Sc. degree in mechanical engineering (computational mechanics) and a M.Sc. degree in civil engineering from Federal University of Pernambuco, Brazil in 2011 and 2014, respectively. He is currently a Ph.D. candidate in the Energy Resources Engineering department at Stanford University where he is investigating (under a SwB fellowship) stochastic models of landscape evolution. His research interests include geostatistics, machine learning, convex optimization, computer vision, and high-performance computing.
\end{IEEEbiography}

\begin{IEEEbiography}[{\includegraphics[width=1in,height=1.25in,clip,keepaspectratio]{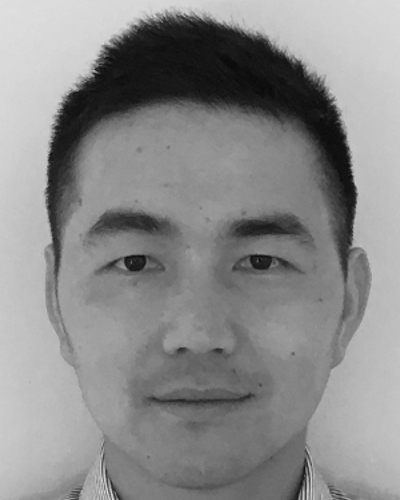}}]{Youli Mao}
received a BS degree in applied mathematics from Sichuan University, China, in 2009 and a PhD degree in applied mathematics from Texas A\&M University in 2014. He is currently a graduate student in the Geology and Geophysics department at Texas A\&M University. His research interests include the integration of fluid mechanics and geomechanics, machine learning in geophysics and data science.
\end{IEEEbiography}


\begin{IEEEbiography}[{\includegraphics[width=1in,height=1.25in,clip,keepaspectratio]{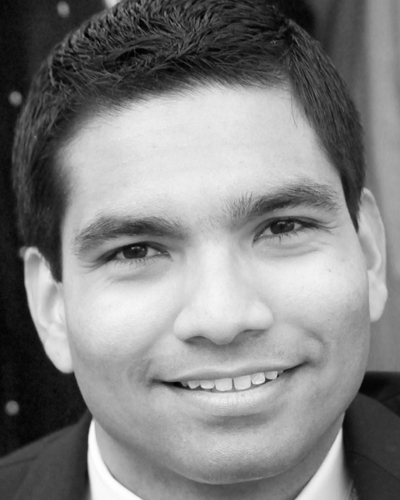}}]{Avinash Wesley}
received a BS degree in computer science from St. Stephen's College, Delhi India, in 2006, a MS degree in computer science from the University of Houston in 2010, and a PhD degree in computer science from the University of Houston in 2015. He is currently a senior technologist at Halliburton. His interests are in data science, IoT, and deep learning. He also develops real-time solutions for drilling and production operations in the oil and gas industry.
\end{IEEEbiography}

\begin{IEEEbiography}[{\includegraphics[width=1in,height=1.25in,clip,keepaspectratio]{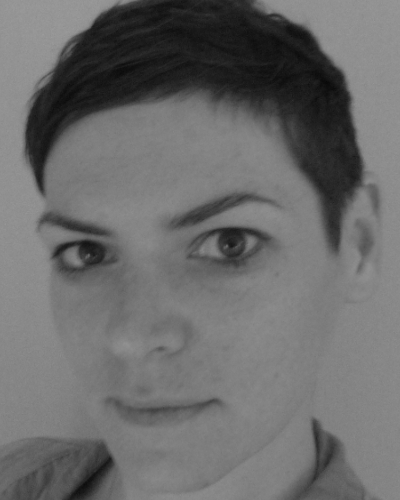}}]{Aimee Taylor}
received a BA degree in economics from University of Leeds and a BSc degree in geology from Durham University. She is currently a senior product specialist for real-time analytics and predictive analytics at Halliburton, focusing on developing automated processes to facilitate and streamline operations.
\end{IEEEbiography}


\vfill


\end{document}